\documentclass[preprint,3p,times,twocolumn]{elsarticle}

\usepackage{amssymb}

\usepackage{float}
\usepackage{stfloats}

\usepackage{booktabs}

\journal{Image and Vision Computing}

\begin{document}

\begin{frontmatter}



\title{From Pixels to Sentiment: Fine-tuning CNNs for Visual Sentiment Prediction}


\author[bsc]{V{\'i}ctor Campos}
\ead{victor.campos@bsc.es}
\author[columbia]{Brendan Jou}
\ead{bjou@caa.columbia.edu}
\author[upc]{Xavier Gir{\'o}-i-Nieto}
\ead{xavier.giro@upc.edu}

\address[bsc]{Barcelona Supercomputing Center (BSC), Barcelona, Catalonia/Spain}
\address[columbia]{Columbia University, New York, NY USA}
\address[upc]{Universitat Polit{\`e}cnica de Catalunya (UPC), Barcelona, Catalonia/Spain}


\begin{abstract}
Visual multimedia have become an inseparable part of our digital social lives, and they often capture moments tied with deep affections. Automated visual sentiment analysis tools can provide a means of extracting the rich feelings and latent dispositions embedded in these media. In this work, we explore how Convolutional Neural Networks (CNNs), a now de facto computational machine learning tool particularly in the area of Computer Vision, can be specifically applied to the task of visual sentiment prediction. We accomplish this through fine-tuning experiments using a state-of-the-art CNN and via rigorous architecture analysis, we present several modifications that lead to accuracy improvements over prior art on a dataset of images from a popular social media platform. We additionally present visualizations of local patterns that the network learned to associate with image sentiment for insight into how visual positivity (or negativity) is perceived by the model. 
\end{abstract}

\begin{keyword}Sentiment \sep Convolutional Neural Networks \sep Social Multimedia \sep Fine-tuning Strategies



\end{keyword}

\end{frontmatter}


\section{Introduction}
\label{introduction}
The shear throughput of user-generated multimedia content uploaded to social networks every day has experienced tremendous growth in the last several years. These social networks often serve as platforms for their users to express feelings and opinions. And visual multimedia, in particular, has become a natural and rich form to communicate emotions and sentiments in a host of these digital media platforms.

Affective Computing \cite{picard_1997} is lately drawing increased attention by multiple research disciplines. This increased interest may be attributed to recent successes in areas like emotional understanding of viewer responses to advertisements using facial expressions \cite{mcduffpredicting} and monitoring of emotional patterns to help patients suffering from mental health disorder \cite{huang_2014}. Given the complexity of the task, visual understanding for emotion and sentiment detection has lagged behind other Computer Vision tasks, e.g., in general object recognition.

\emph{Emotion} and \emph{sentiment} are closely connected entities. Emotion is usually defined as high intensity, but relatively brief experience, onset by a stimuli \cite{plutchik_1980, cabanac_2002}, whereas sentiment refers to an attitude, disposition or opinion towards a certain topic \cite{pang_2008} and usually implies a longer-lived phenomena than that in emotion. Throughout this work we represent sentiment values as a polarity that can be either \emph{positive} or \emph{negative}, although some works also consider the \emph{neutral} class or even a finer scale that accounts for different strengths \cite{xu2014visual}. Since the data used in our experiments is annotated using crowdsourcing, we believe that binary binning was helpful to force the annotators to decide between either polarities rather than tend toward a neutral rating. 

The state-of-the-art in classical Computer Vision tasks have recently undergone rapid transformations thanks to the re-popularization of Convolutional Neural Networks (CNNs) 
\cite{krizhevsky_2012,chen_2014}. This led us to also explore such architectures for visual sentiment prediction where we seek to recognize the sentiment 
that an image would provoke to a human viewer. Given the challenge of collecting large-scale datasets with reliable sentiment annotations, our efforts focus on understanding domain-transferred CNNs for visual sentiment prediction by analyzing the performance of a state-of-the-art architecture fine-tuned for this task.

In this paper, we extend our previous work in \cite{campos_2015}, where we empirically studied the suitability of domain transferred CNNs for visual sentiment prediction. The new contributions of this paper include: (1) an extension of the fine-tuning experiment on a larger set of images with more ambiguous annotations, (2) a study of the impact of weight initialization by varying the source domain from which we transfer learning from, (3) an improved network architecture based on empirical insights, and (4) a visualization of the local image regions that contribute to the overall sentiment prediction.


\begin{figure}
		\includegraphics[width=\linewidth]{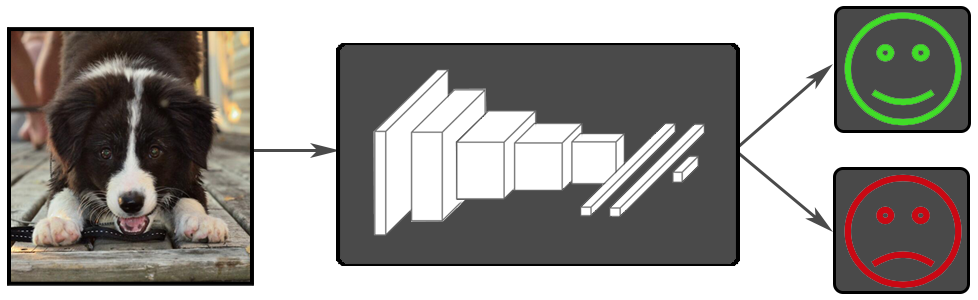}
		\caption{Overview of the proposed visual sentiment prediction framework.} 
		\label{fig:overview}
\end{figure}

\section{Related Work}
\label{related_work}
Computational affective understanding for visual multimedia is a growing area of research interest and historically has benefited from application of a classical handcrafted vision feature representations. For example, color histograms and SIFT-based Bag-of-Words, hallmark low-level image descriptors, were applied in \cite{siersdorfer_2010} for visual sentiment prediction. Likewise, art- and psychology-inspired visual descriptors were used in visual emotion classification \cite{machajdik_2010} and automatic image emotion adjustment \cite{peng_2015}. In \cite{borth_2013} and \cite{jou_2015}, visual sentiment ontologies consisting of adjective-noun pairs (ANPs) were proposed as a mid-level representation for bridging the \emph{affective gap} between low-level visual features and high-level affective semantics. A bank of detectors was also proposed in \cite{borth_2013} and \cite{jou_2015}, referred to as SentiBank and MVSO, respectively, to detect these mid-level representations in input images and use them in visual sentiment prediction tasks. Unlike some of these methods, which are trained and evaluated on datasets with weak labels mined from data, our work focuses on images with crowdsourced sentiment labels.

Convolutional Neural Networks (CNNs) \cite{lecun_1998} are enjoying enormous research attention in recent Computer Vision research. It may be argued that the arrival of large-scale datasets like \cite{deng_2009} and the democratization of graphical processing units (GPUs) has led CNNs to the outstanding vision successes they have experienced, e.g., \cite{krizhevsky_2012,he_2015,szegedy_2014}. In application to areas where large-scale data are much more difficult to gather, CNNs have still proven effective through the use of transfer learning \cite{oquab_2014}. In such transfer learning settings, pre-trained CNNs are used either as off-the-shelf feature extractors where embeddings are taken from intermediate layers activations \cite{donahue_2014,razavian_2014} or as weight initializers for fine-tuning to the new target task \cite{salvador_2015}. In general, standard fine-tuning, e.g., as in \cite{girshick_2014}, have shown superior performance as compared to using CNNs just as generic feature extractors \cite{agrawal_2014}, albeit coming at the cost of additional training. Further insights on the best practices for the fine-tuning were also developed in \cite{chubest_2016}, where the suggestions were largely domain-specific and depending on the visual similarities between source and target domains.


Recent work in applying CNNs to visual sentiment transfer learning was explored in \cite{xu2014visual}, where it was shown that off-the-shelf visual descriptors could outperform hand-crafted low-level features and SentiBank \cite{borth_2013}. The application of CNNs for visual sentiment prediction was further explored in \cite{you_2015}, where a CNN was developed for such task, but little intuition for why their network would improve on the state-of-the-art architectures was given. In this work, we pre-train with a classical, but proven CNN model and develop a thorough analysis of the network in order to gain insight in the design and training of CNNs for the task of visual sentiment prediction.

\section{Methodology}
\label{methodology}
\begin{figure}[t]
		\centering
		\includegraphics[width=0.7\linewidth]{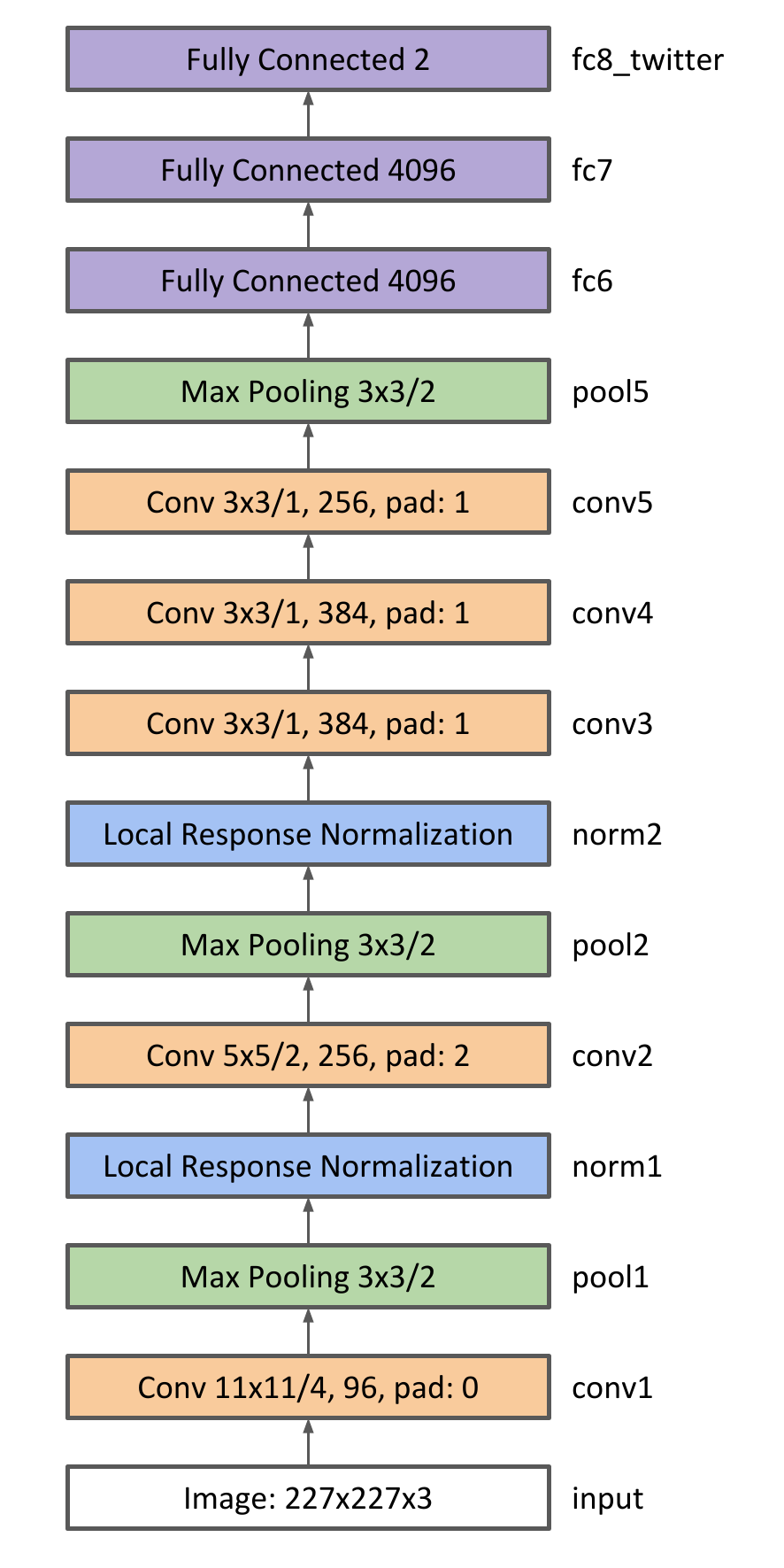}
		\caption{The template Convolutional Neural Network architecture employed in our experiments, an AlexNet-styled architecture \cite{krizhevsky_2012} adapted for visual sentiment prediction.} 
		\label{fig:caffenet_architecture}
\end{figure}


In this work, we used the CaffeNet CNN architecture \cite{jia_2014}, an AlexNet-styled network that differs from the ILSVRC2012 winning architecture \cite{krizhevsky_2012} in the order of the pooling and normalization layers. As depicted in Figure \ref{fig:caffenet_architecture}, the architecture is composed of five convolutional layers and three fully-connected layers. Rectified linear unit (ReLU) non-linearities, $\max(0,\mathbf{x})$, are used as the activations throughout the network. The first two convolutional layers are followed by max pooling and local response normalization (LRN), and the fifth convolutional layer \emph{conv5} is followed by max pooling. The output of the last fully-connected layer \emph{fc8} is fed to a softmax that computes the probability distribution over the target classes. Our experiments were performed using Caffe \cite{jia_2014}, a publicly available deep learning framework.

In this work, we used the Twitter dataset collected and released in \cite{you_2015}, also called DeepSent, to train and evaluate our fine-tuned networks for visual sentiment prediction. In contrast with many other annotation approaches which rely on image metadata, usually producing weak labels, each of the 1269 images in the dataset were labeled for either positive or negative sentiment by five human annotators. This annotation process was carried out using the Amazon Mechanical Turk crowdsourcing platform (for more details on the dataset construction, please see \cite{you_2015}). 
We use the subset of images where there was a consensus across all five annotators, also called \emph{five-agree subset} in \cite{you_2015}. The 880 images in the \emph{five-agree subset} were divided into five different folds to obtain more statistically meaningful results by applying cross-validation.

\subsection{Fine-tuning CaffeNet for Visual Sentiment}
\label{section:meth-finetuning_caffenet}
Convolutional Neural Networks (CNNs) often contain a large number of parameters that need to be learned, and also often require large datasets when training from scratch. In visual sentiment prediction tasks though, the size of the datasets is usually constrained due to the difficulty and expense of acquiring labels that depend so much on subjective reasoning. A common approach to this problem of small data size is to use transfer learning using information from a pre-trained network trained on a large amount of data to bootstrap the smaller dataset.

In our target setting with the Twitter DeepSent dataset, the number of images available is not large enough to train the some 60 million parameters in CaffeNet from scratch. Fine-tuning is a straightforward transfer learning method applied successfully in previous works \cite{oquab_2014,salvador_2015,agrawal_2014}. Fine-tuning consists of initializing all the weights in the network, except those in the last layer(s), using a pre-trained model instead of random initialization. The last layer is replaced by a new one, usually containing the same amount of neurons as classes in the dataset, with randomly initialized weights. Training then proceeds using the data from the target dataset. The main advantages of this approach are (1) faster convergence, since the gradient descent algorithm starts from a point which is likely much closer to a local minimum, and (2) reduced likelihood of overfitting given the training dataset is small \cite{zeiler_2014, yosinski_2014}. Additionally, in transfer learning settings where the original and target domains are similar, pre-training can be seen as adding additional training data encoded in the pre-trained network. In previous works, AlexNet-styled networks trained on the ILSVRC2012 dataset have proved to learn generic features that perform well in several recognition tasks \cite{donahue_2014, razavian_2014}, and so we use a similar architecture called CaffeNet pre-trained on ILSVRC2012 to perform our fine-tuning.

As shown in Figure \ref{fig:caffenet_architecture}, the original fully-connected \emph{fc8} layer from CaffeNet is replaced by a two-neuron layer, \emph{fc8\_twitter}, representing \emph{positive} and \emph{negative} sentiment. The weights in this new layer are initialized from a zero-mean Gaussian distribution with standard deviation 0.01 and zero bias. The rest of layers are initialized using weights from the pre-trained model. The network is trained using stochastic gradient descent with momentum of 0.9 and an starting learning rate of 0.001 which we decay by a factor of 10 every 6 epochs. Since the last layer was randomly initialized rather than pre-trained, its learning rate was set 10 times higher than the base. Each model is trained for 65 epochs using mini-batches of 256 images.

One simple technique that has proven quite effective in tasks like object recognition \cite{chatfield_2014} is oversampling, which consists of feeding slightly modified versions of the image (e.g.,~by applying flips and crops) to the network during test time and averaging prediction results. This serves as a type of model ensembling and helps to deal with the dataset bias \cite{torralba_2011}. We also use oversampling in our visual sentiment prediction setting by feeding 10 combinations of flips and crops of the original image to the CNN during test.


\subsection{Layer-wise Analysis}

\begin{figure}
		\includegraphics[width=\linewidth]{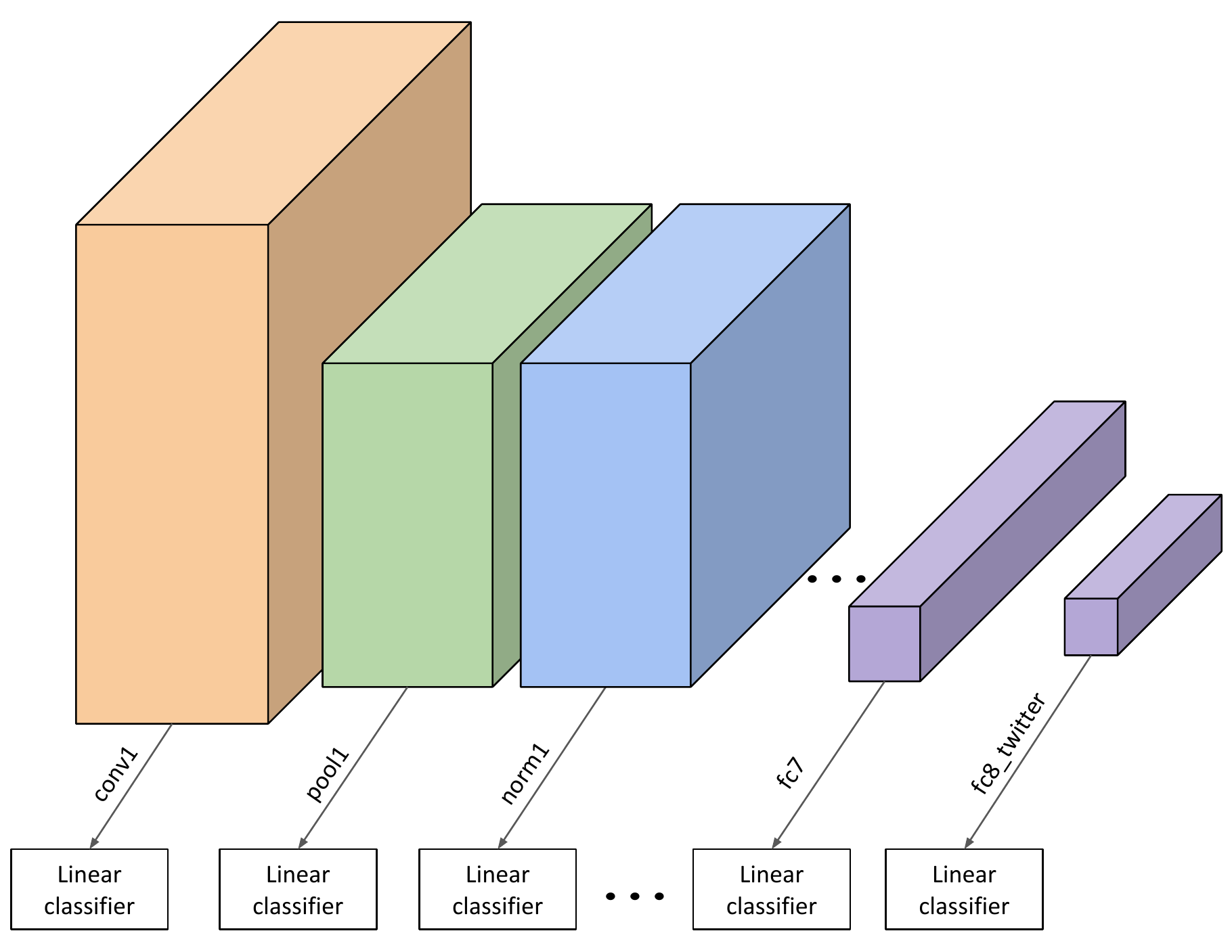}
		\caption{Experimental setup for the layer analysis using linear classifiers. Activations in each layer are used as visual descriptors in order to train a classifier.}
		\label{fig:linear_classfiers}
\end{figure}

In Section \ref{section:res-linear_classifiers}, we present a series of experiments to analyze the contribution of individual layers in our fine-tuning of the CaffeNet architecture for visual sentiment prediction.
To accomplish this, we extract the output of weight layers post-activation and use them as visual descriptors.



Previous works have used the activations from individual layers as visual descriptors to solve different vision tasks \cite{salvador_2015,razavian_2014}, although only fully-connected layers are usually used for this purpose. We further extend this idea and train classifiers using activations from all the layers in the architecture, as depicted in Figure \ref{fig:linear_classfiers}, so it is possible to compare the effectiveness of the different representations that are learned along the network. Feature maps from convolutional, pooling and normalization layers were flattened into \emph{d}-dimensional vectors before being used to train the classifiers. Two different classifiers were considered: Support Vector Machine (SVM) with linear kernel and Softmax. The regularization parameter of each classifier was optimized by cross-validation.

\subsection{Layer Ablation}
\label{section:layer_ablation}


\begin{figure}
		\includegraphics[width=\linewidth]{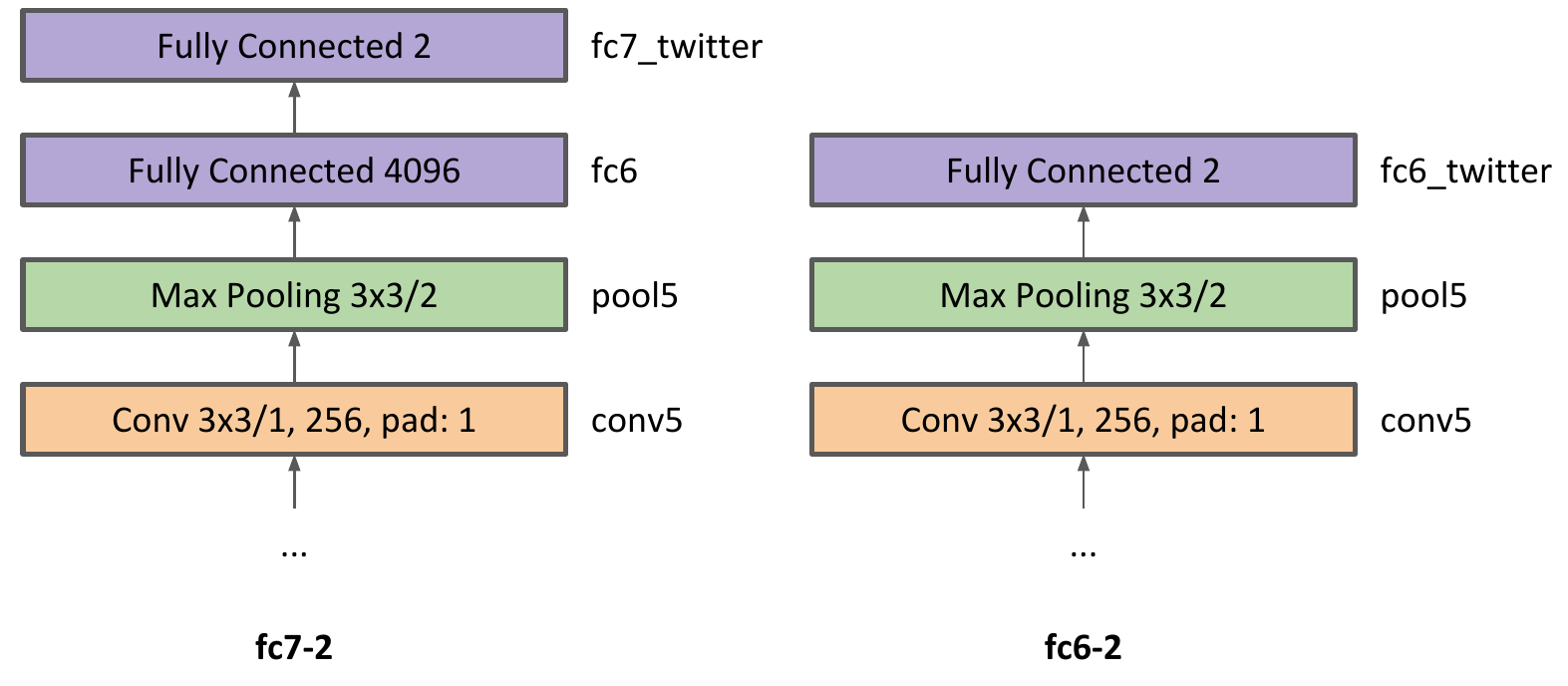}
		\caption{Layer ablation architectures. The 2-neuron layer on top of each architecture is initialized with random weights, whereas the rest of parameters in the network are loaded from the pre-trained model.} 
		\label{fig:layer_ablation_architectures}
\end{figure}

It is not always immediately apparent how much each layer contributes to the ultimate performance of a network. This has leds to analyses and proposed improvements in both CNNs \cite{chatfield_2014,zeiler_2014} and RNNs \cite{graves_2016_act}.
In our experiments presented in \ref{results:ablation}, we show how fully-connected layers, a substantial portion of the network's parameters, affects the performance during CNN fine-tuning for the task of visual sentiment prediction. In particular, the two different architectures in Figure \ref{fig:layer_ablation_architectures} are studied, wherein the last or the two last fully-connected layers are removed, denoted as \emph{fc6-2} and \emph{fc7-2}, respectively.

The last layer always contains as many hidden units as there are classes in the dataset, in this case, just two neurons, one for \emph{positive} and one for \emph{negative} sentiment. 
Weight initialization for the new layers, hyperparameters and training conditions follow the procedure described in Section \ref{section:meth-finetuning_caffenet} except for the learning rate of architecture \emph{fc6-2}.
In practice, for this set of experiments, we found we needed to use a relatively small base learning rate of 0.0001 to get the network to converge.

\subsection{Initialization Analysis}
Since fine-tuning a CNN can be seen as a transfer learning strategy, 
we explored how changing the original domain affects the performance by using different pre-trained models as initialization for the fine-tuning process, while keeping the architecture fixed. In addition to the model trained on ILSVRC 2012 \cite{krizhevsky_2012} (i.e.,~CaffeNet), we evaluate models trained on Places dataset \cite{zhou_2014} (i.e.,~PlacesCNN), which contains images annotated for scene recognition, and two sentiment-related datasets: Visual Sentiment Ontology (VSO) \cite{borth_2013} and Multilingual Visual Sentiment Ontology (MVSO) \cite{jou_2015}, which are used to train adjective-noun pair (ANP) detectors that are later used as a mid-level representations to predict the sentiment in an image. The model trained on VSO, DeepSentiBank \cite{chen_2014}, is a fine-tuning of CaffeNet on VSO. Given the multicultural nature of MVSO, there is one model for each language (i.e.,~English, Spanish, French, Italian, German and Chinese) and each one of them is obtained by fine-tuning DeepSentiBank on a specific language subset of MVSO.
All models are fine-tuned for 65 epochs, following the same procedure as in Section \ref{section:meth-finetuning_caffenet}.

\subsection{Going Deeper: Adding Layers for Fine-tuning}


\begin{figure}[b]
		\includegraphics[width=\linewidth]{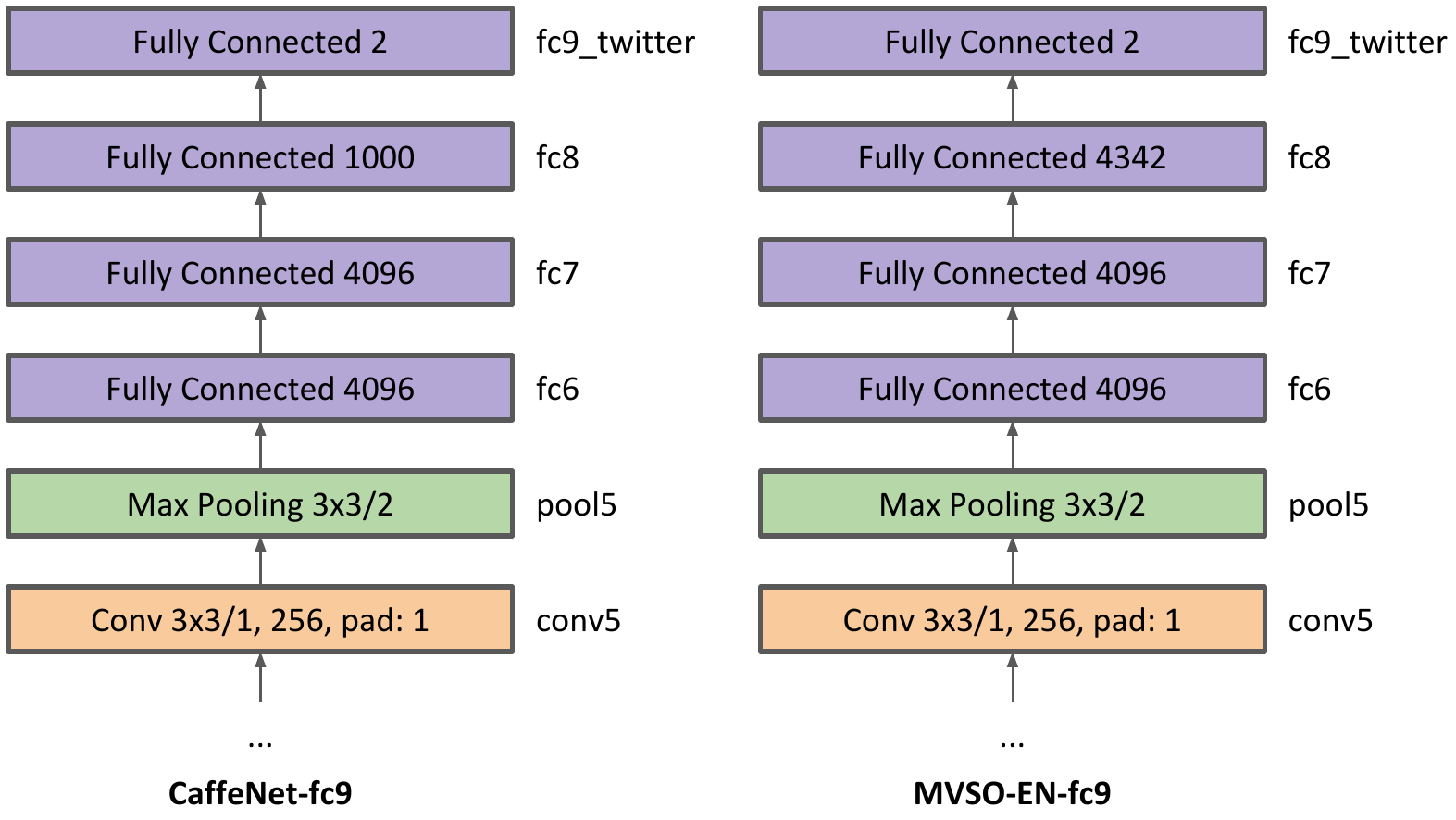}
		\caption{Added fully-connected layers. The whole pre-trained model is loaded and only the new \textit{fc9\_twitter} layer needs to be initialized with random weights.} 
		\label{fig:layer_addition_architectures}
\end{figure}

The activations in a pre-trained CNN's last fully-connected layer contain the likelihood of the input image belonging to each class in the original training dataset, but the regular fine-tuning strategy completely discards this information. Besides, since fully-connected layers contain most of the weights in the architecture, a large amount of parameters that may contain useful information for the target task are being lost.

In this set of experiments, we explore how adding high-level information by reusing the last layer of pre-trained CNNs affects their performance when fine-tuning for visual sentiment prediction. In particular, the networks pre-trained on ILSVRC2012 (i.e.,~CaffeNet) and MVSO-EN are studied. The former was originally trained to recognize 1,000 object classes, whereas the latter was used to detect 4,342 different Adjective Noun Pairs that were designed as a mid-level representation for visual sentiment prediction.

A two-neuron layer, denoted as \emph{fc9\_twitter}, is added on top of both architectures (Figure \ref{fig:layer_addition_architectures}).
We follow the same procedure as described earlier in Section \ref{section:meth-finetuning_caffenet} to initialize the weights in this new layer and train the CNN. The only difference with the previous methodology are the initial values for the weights in the second to last layer, \textit{fc8}, which are now loaded from a pre-trained model instead of being generated following a random probability distribution.

\subsection{Visualization with Fully Convolutional Networks}
One natural approach to gain insight into how concepts are learned by the network is by observing which patches of an image lead the CNN to classify it either as \emph{positive} or \emph{negative}. To do this, we convert our fine-tuned CaffeNet architecture into a fully convolutional network by replacing its fully-connected layers by convolutional layers (see Table \ref{table:fully_conv_layers} for details), following the method in \cite{long_2015} and rearrange the learned model weights from the fully-connected layers into convolutional layers. No additional training is needed for this visualization, only shuffling of weights.

\begin{table}
	\centering
    \label{table:fully_conv_layers}
	\resizebox{\linewidth}{!}{
      \begin{tabular}{cccc}
        \toprule
        \textbf{Layer} & \textbf{Number of kernels} & \textbf{Kernel size ($\mathbf{h \times w \times d}$)}\\
        \midrule
        fc6-conv & 4096 & $6\times 6\times 256$\\
        \midrule
        fc7-conv & 4096 & $1\times 1\times 4096$\\
        \midrule
        fc8\_twitter-conv & 2 & $1\times 1\times 4096$\\
        \bottomrule
      \end{tabular}
	} 
    \caption{Details of new convolutional layers resulting from converting our CaffeNet to a fully convolutional network (stride=1).}
\end{table}

Since the original architecture contains fully-connected layers that implement a dot product operation, it requires the input to have a fixed size. In contrast, the fully convolutional network can handle inputs of any size: by increasing the input size, the dimensions of the output will increase as well and it will become a prediction map on overlapping patches from the input image. We generate $8\times8$ prediction maps for the images of the Twitter five-agree dataset by using inputs of size $451\times451$ instead of $227\times227$, which were the input dimensions of the original architecture.  

\section{Experimental Results}
\label{results}
This section contains the results for the experiments described in Section \ref{methodology}, as well as intuition and conclusions for such results.

\subsection{Fine-tuning CaffeNet for Visual Sentiment}
\label{section:fine-tuning_results}

\begin{table*}
	\centering
    \resizebox{0.7\textwidth}{!}{
      \begin{tabular}{cccc}
        \toprule
        \textbf{Model} & \textbf{Five-agree} & \textbf{Four-agree} & \textbf{Three-agree} \\
        \midrule
        Baseline PCNN from \cite{you_2015} & 0.783 & 0.714 & 0.687\\
        \midrule
        Fine-tuned CaffeNet & 0.817 $\pm$ 0.038 & 0.782 $\pm$ 0.033 & 0.739 $\pm$ 0.033\\
        \midrule
        Fine-tuned CaffeNet with oversampling & \textbf{0.830 $\pm$ 0.034 } & \textbf{0.787 $\pm$ 0.039 } & \textbf{0.749 $\pm$ 0.037 }\\
        \bottomrule
      \end{tabular}
    }
    \caption{Five-fold cross-validation accuracy results on Twitter dataset. Results are displayed as $mean \pm std$.}
    \label{table:fine-tuning_results}
\end{table*}

The five-fold cross-validation results for the fine-tuning experiment on Twitter dataset are detailed in Table \ref{table:fine-tuning_results}, together with the best five-fold cross-validation result in this dataset from \cite{you_2015}. The latter was achieved using a custom architecture, composed by two convolutional layers and four fully-connected layers, that was trained using the Flickr dataset (VSO) \cite{borth_2013} and later fine-tuned on Twitter dataset. In order to evaluate the performance of our approach when using images with more ambiguous annotations, CaffeNet was also fine-tuned on four-agree and three-agree subsets, i.e.,~those containing images that built consensus among at least four and three annotators, respectively.

These results show that, despite being pre-trained for a completely different task, the AlexNet-styled architecture clearly outperforms the custom architecture from \cite{you_2015}. This difference suggests that visual sentiment prediction architectures may benefit from an increased depth that comes from adding a larger amount of convolutional layers instead of fully-connected ones, as suggested by \cite{zeiler_2014} for the task of object recognition. Secondly, these results highlight the importance of high-level representations for the addressed task, as transferring learning from object recognition to sentiment prediction results in high accuracy rates.

Averaging over the predictions of modified versions of the image results in an additional performance boost, as found out by the authors in \cite{chatfield_2014} for the task of object recognition. This fact suggests that oversampling helps to compensate the dataset bias and increases the generalization capability of the system without a penalization on the prediction speed thanks to the batch computation capabilities of GPUs.


\subsection{Layer-wise Analysis}
\label{section:res-linear_classifiers}

The results for the layer-wise analysis using linear classifiers are compared in Table \ref{table:linear_classifiers_results}. The evolution of the accuracy rates at each layer, for both SVM and Softmax classifiers, shows how the learned representation becomes more effective along the network. While every single layer does not introduce a performance boost with respect to the previous ones, it does not necessarily mean that the architecture needs to be modified: since the training of the network is performed in an end-to-end manner, some of the layers may apply a transformation to their inputs from which later layers may benefit, e.g.~\emph{conv5} and \emph{pool5} report lower accuracy than the previous \emph{conv4} when used directly for classification, but the fully-connected layers on top of the architecture may be benefiting from their effect since they produce higher accuracy rates than \emph{conv4}.


Previous works have studied the suitability of Support Vector Machines to classify \emph{off-the-shelf} visual descriptors extracted from pre-trained CNNs \cite{razavian_2014}, while some others have even trained these networks using the L2-SVM's squared hinge loss on top of the architecture \cite{tang_2013}. From our layer-wise analysis, it is not possible to claim that one of the classifiers consistently outperforms the other for the task of visual sentiment prediction, at least using the proposed CNN in the Twitter five-agree dataset.

\begin{table} 
	\centering
    \resizebox{0.65\linewidth}{!}{
      \begin{tabular}{ccc}
        \toprule
        \textbf{Layer}&\textbf{SVM}&\textbf{Softmax}\\
        \midrule
        \emph{fc8} & 0.82 $\pm$ 0.055 & 0.821 $\pm$ 0.046 \\
        \midrule
        \emph{fc7} & 0.814 $\pm$ 0.040 & 0.814 $\pm$ 0.044 \\
        \midrule
        \emph{fc6} & 0.804 $\pm$ 0.031 & 0.81 $\pm$ 0.038 \\
        \midrule
        \emph{pool5} & 0.784 $\pm$ 0.020 & 0.786 $\pm$ 0.022 \\
        \midrule
        \emph{conv5} & 0.776 $\pm$ 0.025 & 0.779 $\pm$ 0.034 \\
        \midrule
        \emph{conv4} & 0.794 $\pm$ 0.026 & 0.781 $\pm$ 0.020 \\
        \midrule
        \emph{conv3} & 0.752 $\pm$ 0.033 & 0.748 $\pm$ 0.029 \\
        \midrule
        \emph{norm2} & 0.735 $\pm$ 0.025 & 0.737 $\pm$ 0.021 \\
        \midrule
        \emph{pool2} & 0.732 $\pm$ 0.019 & 0.729 $\pm$ 0.022 \\
        \midrule
        \emph{conv2} & 0.735 $\pm$ 0.019 & 0.738 $\pm$ 0.030 \\
        \midrule
        \emph{norm1} & 0.706 $\pm$ 0.032 & 0.712 $\pm$ 0.031 \\
        \midrule
        \emph{pool1} & 0.674 $\pm$ 0.045 & 0.68 $\pm$ 0.035 \\
        \midrule
        \emph{conv1} & 0.667 $\pm$ 0.049 & 0.67 $\pm$ 0.032 \\
        \bottomrule
      \end{tabular}
    }
    \caption{Layer analysis with linear classifiers. Results are given in $mean \pm std$ five-fold cross-validation accuracy on the five-agree DeepSent Twitter dataset.}
    \label{table:linear_classifiers_results}
\end{table}


\subsection{Layer Ablation}
\label{results:ablation}
The five-fold cross-validation results for the fine-tuning of the ablated architectures are shown in Table \ref{table:layer_removal_results}. Following the behavior observed in the layer-wise analysis with linear classifiers in Section \ref{section:res-linear_classifiers}, removing layers from the top of the architecture results in a deterioration of the classification accuracy.


The drop in accuracy for architecture \emph{fc6-2} is larger than one may expect given the results from the layer by layer analysis, which denotes that the convergence from 9,216 neurons in \emph{pool5} to a two-layer neuron might be too sudden. This is not the case of architecture \emph{fc7-2}, where the removal of more than 16M parameters produces only a slight deterioration in performance. These observations suggest that an intermediate fully-connected layer that provides a softer dimensionality reduction is beneficial for the architecture, but the addition of a second fully-connected layer between \emph{pool5} and the final two-neuron layer produces a small gain compared to the extra 16M parameters that are being added. This trade-off is especially important for tasks such as visual sentiment prediction, where collecting large datasets with reliable annotations is difficult, and removing one of the fully-connected layers in the architecture might allow training it from scratch using smaller datasets without overfitting the model.

\begin{table*}[t]
	\centering
	\resizebox{0.7\textwidth}{!}{
      \begin{tabular}{cccc}
        \toprule
        \textbf{Architecture}&\textbf{Without oversampling}&\textbf{With oversampling}&\textbf{Parameter reduction}\\
        \midrule
        fc7-2 & 0.784 $\pm$ 0.024 & 0.797 $\pm$ 0.021 & $>$16M\\
        \midrule
        fc6-2 & 0.651 $\pm$ 0.044 & 0.676 $\pm$ 0.029 & $>$54M\\
        \bottomrule
      \end{tabular}
	} 
    \caption{Layer ablation: Five-fold cross-validation accuracy results on five-agree Twitter dataset. Results are displayed as $mean \pm std$.}
    \label{table:layer_removal_results}
\end{table*}

\subsection{Initialization Analysis}
Convolutional Neural Networks trained from scratch using large-scale datasets usually achieve very similar results regardless of their initialization, however, for our visual sentiment prediction task, fine-tuning on a smaller dataset using different weight initialization under low learning rate conditions does seem to variably influence the final performance. This is shown by the results for the different initializations in Table \ref{table:initialization_results}.

These empirical results show how most of the models that were already trained for a sentiment-related task outperform the ones pre-trained on ILSVRC 2012 and Places, whose images are mostly neutral in terms of sentiment. 
Because the Twitter dataset used in our experiments was labeled using Amazon Mechanical Turk, the annotators were required to be U.S.~residents, introducing a certain culture bias in the annotations. This, together with the performance gap observed with the MVSO-ZH model compared to the rest of MVSO models, suggests the potential of an image sentiment perception gap between Eastern and Western cultures. A similar behavior was observed in \cite{jou_2015}, where the authors reported that using a Chinese-specific model to predict the sentiment in other languages reported the worst results in all their cross-lingual domain transfer experiments.

\begin{figure}[b]
		\includegraphics[width=\linewidth]{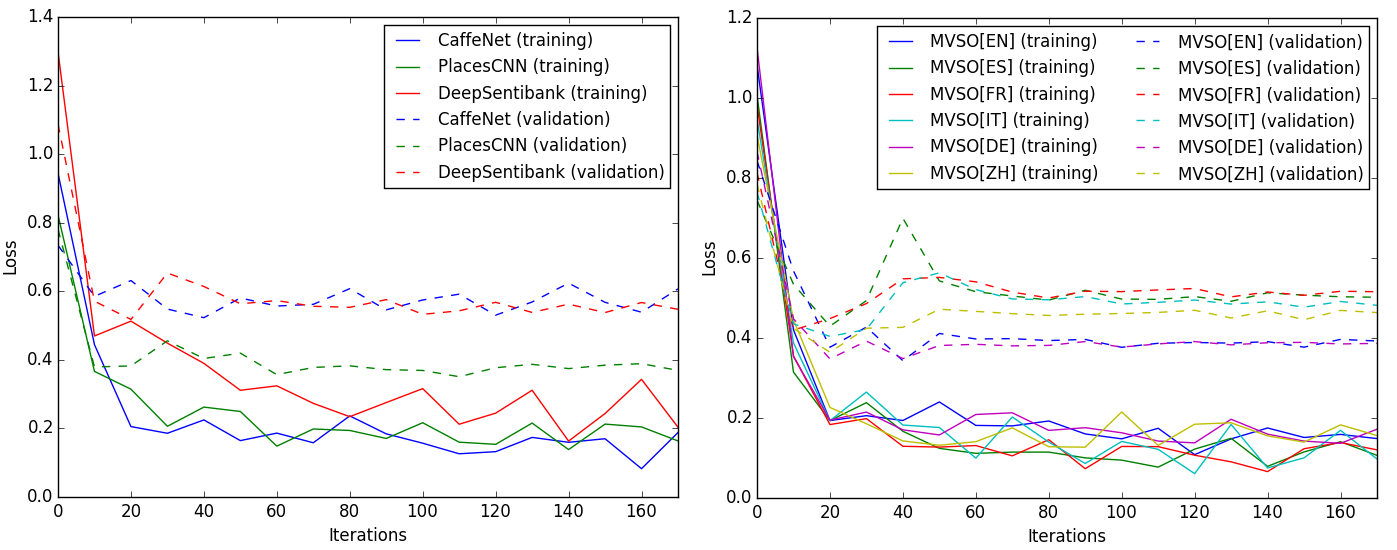}
		\caption{Comparison of the evolution of the loss function on one of the folds during training.}
		\label{fig:loss_graph}
\end{figure}

A comparison of the evolution of the loss function of the different models during training can be seen in Figure \ref{fig:loss_graph}, where it can be observed that the different pre-trained models need a different amount of iterations until convergence. The DeepSentiBank model seems to adapt worse than other models to the target dataset albeit being pre-trained for a sentiment-related task, as can be seen both in its final accuracy and in its noisy and slow evolution during training. On the other hand, the different MVSO models not only provide the top accuracy rates, but converge faster and in a smoother way as well.

\begin{table}
	\centering
    \resizebox{\linewidth}{!}{
      \begin{tabular}{ccc}
        \toprule
        \textbf{Pre-trained model}&\textbf{Without oversampling}&\textbf{With oversampling}\\
        \midrule
        CaffeNet & 0.817 $\pm$ 0.038 & 0.830 $\pm$ 0.034 \\
        \midrule
        PlacesCNN & 0.823 $\pm$ 0.025 & 0.823 $\pm$ 0.026 \\
        \midrule
        DeepSentiBank & 0.804 $\pm$ 0.019 & 0.806 $\pm$ 0.019 \\
        \midrule
        MVSO [EN] & \textbf{0.839 $\pm$ 0.029} & \textbf{0.844 $\pm$ 0.026} \\
        \midrule
        MVSO [ES] & 0.833 $\pm$ 0.024 & \textbf{0.844 $\pm$ 0.026} \\
        \midrule
        MVSO [FR] & 0.825 $\pm$ 0.019 & 0.828 $\pm$ 0.012 \\
        \midrule
        MVSO [IT] & 0.838 $\pm$ 0.020 & 0.838 $\pm$ 0.012 \\
        \midrule
        MVSO [DE] & 0.837 $\pm$ 0.025 & 0.837 $\pm$ 0.033 \\
        \midrule
        MVSO [ZH] & 0.797 $\pm$ 0.024 & 0.806 $\pm$ 0.020 \\
        \bottomrule
      \end{tabular}
    }
    \caption{Five-fold cross-validation $mean \pm std$ accuracies for different network weight initialization schemes on the five-agree DeepSent Twitter dataset.}
    \label{table:initialization_results}
\end{table}

\begin{figure*}[t]
  \centering
  \includegraphics[width=\textwidth]{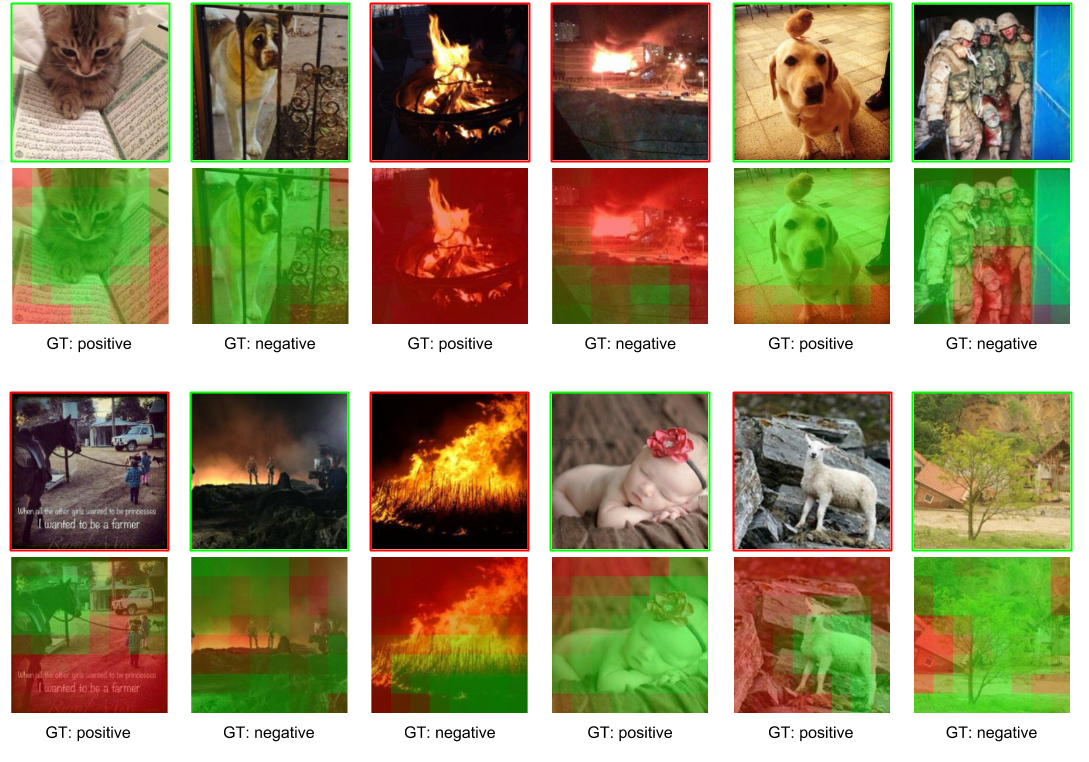}
  \caption{Some examples of the global and local sentiment predictions of the fine-tuned MVSO-EN CNN. The color of the border indicates the predicted sentiment at global scale, i.e.,~green for \emph{positive} and red for \emph{negative}. The heatmaps in the second row follow the same color code, but they are not binary: a higher intensity means a stronger prediction towards the represented sentiment.}
  \label{fig:visualization_examples}
\end{figure*}

\subsection{Going Deeper: Adding Layers for Fine-tuning}

\begin{table}
	\centering
	\resizebox{\linewidth}{!}{
      \begin{tabular}{cccc}
        \toprule
        \textbf{Architecture}&\textbf{Without oversampling}&\textbf{With oversampling}\\
        \midrule
        CaffeNet-fc9 & 0.795 $\pm$ 0.023 & 0.803 $\pm$ 0.034\\
        \midrule
        MVSO-EN-fc9 & 0.702 $\pm$ 0.067 & 0.694 $\pm$ 0.060\\
        \bottomrule
      \end{tabular}
	} 
    \caption{Adding Layers: Five-fold cross-validation accuracy results on five-agree Twitter dataset. Results are displayed as $mean \pm std$.}
    \label{table:layer_addition_results}
\end{table}

The results for the layer addition experiments, which are compared in Table \ref{table:layer_addition_results}, show that the accuracy achieved by reusing all the information in the original models is poorer than when performing a regular fine-tuning.

One possible reason for the loss of performance with respect to the regular fine-tuning is the actual information being reused by the network. For instance, the CaffeNet model was trained on ILSVRC 2012 for the recognition of objects which are mostly neutral in terms of sentiment, e.g.~\emph{teapot}, \emph{ping-pong ball} or \emph{apron}. This is not the case of MVSO-EN, which was originally used to detect sentiment-related concepts such as \emph{nice car} or \emph{dried grass}. The low accuracy rates of this last model may be justified by the low ANP detection rate of the original MVSO-EN model (0.101 top-1 ANP detection accuracy in a classification task with 4,342 classes), as well as by a mismatch between the concepts in the original and target domains.

Moreover, the MVSO-EN CNN was originally designed as a mid-level representation, i.e.,~a concept detector that serves as input to a sentiment classifier. This is not being fulfilled when fine-tuning all the weights in the network, so we speculate that freezing the pre-trained layers and learning only the new weights introduced by \emph{fc9\_twitter} may result in a better use of the concept detector and, thus, a boost in performance.

\subsection{Visualization}

Some examples of the visualization results obtained using the fine-tuned MVSO-EN CNN, which is the top performing model among all that have been presented in this work, are depicted in Figure \ref{fig:visualization_examples}. They were obtained by resizing the $8\times8$ prediction maps in the output of the fully convolutional network to fit each image's dimensions. Nearest-neighbor interpolation was used in the resizing process, so that the original prediction blocks were not blurred. The probability for each sentiment, originally in the range $\left[0,1\right]$, was scaled to the range $\left[0, 255\right]$ and assigned to one RGB channel, i.e.~green for \emph{positive} and red for \emph{negative}. It is important to notice that this process is equivalent to feeding 64 overlapped patches of the image to the regular CNN and then composing their outputs to build an $8\times8$ prediction map, but in a much more efficient manner (while the output dimension is 64 times larger, the inference time grows only by a factor of 3). As a consequence, the global prediction by the regular CNN is not the average of the 64 local predictions in the heatmap, but it is still a very useful method to understand the concepts that the model associates to each sentiment.

From the observation of both global and local predictions, we observe two sources of errors that may be addressed in future experiments. Firstly, a lack of granularity in the detection of some high-level semantics is detected, e.g.~the network seems unable to tell a campfire from a burning building, and associates them to the same sentiment. On the other hand, the decision seems to be driven mainly by the main object or concept in the image, whereas the context is vital for the addressed task. The former source of confusion may be addressed in future research by using larger datasets, while the latter may be improved by using other types of neural networks that have showed increased accuracy in image classification benchmarks, e.g.~Inception \cite{szegedy_2014_googlenet} or ResNet \cite{he_2015_resnet} architectures, or using mid-level representations instead of an end-to-end prediction, e.g.~freezing all the weights in the MVSO models and training just the new \emph{fc9\_twitter} on top of them.

\section{Conclusions and Future Work}
\label{conclusions}
In this work, we have presented extensive experiments comparing several fine-tuned CNNs for visual sentiment prediction. We showed that deep architectures can learn features useful for recognizing visual sentiment in social images, and in particular, several models that outperform the current state-of-the-art on a dataset of Twitter photos was presented. Some of these models outperform the state-of-the-art with a smaller number of parameters compared to the original architecture. These observations along with others have highlighted the importance of empirical insights and guided sweeps over the space of network designs. We also showed that the choice of pre-training in model initialization can indeed make a difference when the target dataset is small. In addition, to better understand these models, we presented a sentiment prediction visualization with spatial localization that helped further insights into erroneous classifications as well as better understand learned network representations.

In the future, we plan to study other state-of-the-art convolutional network architectures for visual sentiment analysis. In addition, we will seek to expand our analysis to larger scale and weakly supervised settings as well as develop models that can learn reliably under noisy label conditions.

\section*{Acknowledgments}
\label{acknowledgements}

This work has been developed in the framework of the BigGraph TEC2013-43935-R project, funded by the Spanish Ministerio de Econom\'ia y Competitividad and the European Regional Development Fund (ERDF). It has been supported by the Severo Ochoa Program's SEV2015-0493 grant awarded by the Spanish Government, the TIN2015-65316 project by the Spanish Ministerio de Econom\'ia y Competitividad and contracts 2014-SGR-1051 by Generalitat de Catalunya. The Image Processing Group at the UPC is a SGR14 Consolidated Research Group recognized and sponsored by the Catalan Government (Generalitat de Catalunya) through its AGAUR office. We gratefully acknowledge the support of NVIDIA Corporation with the donation of the GeForce GTX Titan Z and X used in this work and the support of BSC/UPC NVIDIA GPU Center of Excellence. 

\section*{References}
\bibliographystyle{elsarticle-num} 
\bibliography{sigproc} 

\begin{thebibliography}{10}
\expandafter\ifx\csname url\endcsname\relax
  \def\url#1{\texttt{#1}}\fi
\expandafter\ifx\csname urlprefix\endcsname\relax\def\urlprefix{URL }\fi
\expandafter\ifx\csname href\endcsname\relax
  \def\href#1#2{#2} \def\path#1{#1}\fi

\bibitem{picard_1997}
R.~W. Picard, Affective Computing, Vol. 252, MIT Press Cambridge, 1997.

\bibitem{mcduffpredicting}
D.~McDuff, R.~El~Kaliouby, J.~F. Cohn, R.~W. Picard, Predicting ad liking and
  purchase intent: {L}arge-scale analysis of facial responses to ads, 2015.

\bibitem{huang_2014}
S.~T.-Y. Huang, A.~Sano, C.~M.~Y. Kwan, The moment: {A} mobile tool for people
  with depression or bipolar disorder, in: ACM International Joint Conference
  on Pervasive and Ubiquitous Computing: Adjunct Publication, 2014.

\bibitem{plutchik_1980}
R.~Plutchik, Emotion: {A} Psychoevolutionary Synthesis, Harper \& Row, 1980.

\bibitem{cabanac_2002}
M.~Cabanac, What is emotion?, Behavioural processes 60~(2) (2002) 69--83.

\bibitem{pang_2008}
B.~Pang, L.~Lee, Opinion mining and sentiment analysis, Information Retrieval
  2~(1-2) (2008) 1--135.

\bibitem{xu2014visual}
C.~Xu, S.~Cetintas, K.-C. Lee, L.-J. Li, Visual sentiment prediction with deep
  convolutional neural networks, 2014.

\bibitem{krizhevsky_2012}
A.~Krizhevsky, I.~Sutskever, G.~E. Hinton, Image{N}et classification with deep
  convolutional neural networks, in: Advances in Neural Information Processing
  Systems (NIPS), 2012.

\bibitem{chen_2014}
T.~Chen, D.~Borth, T.~Darrell, S.-F. Chang, {DeepSentiBank}: {V}isual sentiment
  concept classification with deep convolutional neural networks, 2014.

\bibitem{campos_2015}
V.~Campos, A.~Salvador, B.~Jou, X.~Giro-i Nieto, Diving deep into sentiment:
  {U}nderstanding fine-tuned {CNN}s for visual sentiment prediction, in: Intl
  Workshop on Affect \& Sentiment in Multimedia, ACM, 2015.

\bibitem{siersdorfer_2010}
S.~Siersdorfer, E.~Minack, F.~Deng, J.~Hare, Analyzing and predicting sentiment
  of images on the social web, in: ACM Conference on Multimedia (MM), 2010.

\bibitem{machajdik_2010}
J.~Machajdik, A.~Hanbury, Affective image classification using features
  inspired by psychology and art theory, in: ACM Conference on Multimedia (MM),
  2010.

\bibitem{peng_2015}
K.-C. Peng, T.~Chen, A.~Sadovnik, A.~Gallagher, A mixed bag of emotions:
  {M}odel, predict, and transfer emotion distributions, in: IEEE Conference on
  Computer Vision and Pattern Recognition (CVPR), 2015.

\bibitem{borth_2013}
D.~Borth, R.~Ji, T.~Chen, T.~Breuel, S.-F. Chang, Large-scale visual sentiment
  ontology and detectors using adjective noun pairs, in: ACM Conference on
  Multimedia (MM), 2013.

\bibitem{jou_2015}
B.~Jou, T.~Chen, N.~Pappas, M.~Redi, M.~Topkara, S.-F. Chang, Visual affect
  around the world: {A} large-scale multilingual visual sentiment ontology, in:
  ACM Conference on Multimedia (MM), 2015.

\bibitem{lecun_1998}
Y.~LeCun, L.~Bottou, Y.~Bengio, P.~Haffner, Gradient-based learning applied to
  document recognition, in: Proceedings of the IEEE, 1998.

\bibitem{deng_2009}
J.~Deng, W.~Dong, R.~Socher, L.-J. Li, K.~Li, L.~Fei-Fei, Image{N}et: {A}
  large-scale hierarchical image database, in: IEEE Conference on Computer
  Vision and Pattern Recognition (CVPR), 2009.

\bibitem{he_2015}
K.~He, X.~Zhang, S.~Ren, J.~Sun, Delving deep into rectifiers: {S}urpassing
  human-level performance on imagenet classification, in: IEEE International
  Conference on Computer Vision (ICCV), 2015.

\bibitem{szegedy_2014}
C.~Szegedy, W.~Zaremba, I.~Sutskever, J.~Bruna, D.~Erhan, I.~Goodfellow,
  R.~Fergus, Intriguing properties of neural networks, in: International
  Conference on Learning Representations (ICLR), 2014.

\bibitem{oquab_2014}
M.~Oquab, L.~Bottou, I.~Laptev, J.~Sivic, Learning and transferring mid-level
  image representations using convolutional neural networks, in: IEEE
  Conference on Computer Vision and Pattern Recognition (CVPR), 2014.

\bibitem{donahue_2014}
J.~Donahue, Y.~Jia, O.~Vinyals, J.~Hoffman, N.~Zhang, E.~Tzeng, T.~Darrell,
  {DeCAF}: {A} deep convolutional activation feature for generic visual
  recognition., in: International Conference on Machine Learning (ICML), 2014.

\bibitem{razavian_2014}
A.~S. Razavian, H.~Azizpour, J.~Sullivan, S.~Carlsson, {CNN} features
  off-the-shelf: {A}n astounding baseline for recognition, in: IEEE Conference
  on Computer Vision and Pattern Recognition Workshops (CVPRW), 2014.

\bibitem{salvador_2015}
A.~Salvador, M.~Zeppelzauer, D.~Manchon-Vizuete, A.~Calafell, X.~Giro-i Nieto,
  Cultural event recognition with visual convnets and temporal models, in: IEEE
  Conference on Computer Vision and Pattern Recognition Workshops (CVPRW),
  2015.

\bibitem{girshick_2014}
R.~Girshick, J.~Donahue, T.~Darrell, J.~Malik, Rich feature hierarchies for
  accurate object detection and semantic segmentation, in: IEEE Conference on
  Computer Vision and Pattern Recognition (CVPR), 2014.

\bibitem{agrawal_2014}
P.~Agrawal, R.~Girshick, J.~Malik, Analyzing the performance of multilayer
  neural networks for object recognition, in: European Conference on Computer
  Vision (ECCV), 2014.

\bibitem{chubest_2016}
B.~Chu, V.~Madhavan, O.~Beijbom, J.~Hoffman, T.~Darrell, Best practices for
  fine-tuning visual classifiers to new domains, in: European Conference on
  Computer Vision (ECCV), 2016.

\bibitem{you_2015}
Q.~You, J.~Luo, H.~Jin, J.~Yang, Robust image sentiment analysis using
  progressively trained and domain transferred deep networks, in: AAAI
  Conference on Artificial Intelligence, 2015.

\bibitem{jia_2014}
Y.~Jia, E.~Shelhamer, J.~Donahue, S.~Karayev, J.~Long, R.~Girshick,
  S.~Guadarrama, T.~Darrell, Caffe: {C}onvolutional architecture for fast
  feature embedding, in: ACM Conference on Multimedia (MM), 2014.

\bibitem{zeiler_2014}
M.~D. Zeiler, R.~Fergus, Visualizing and understanding convolutional networks,
  in: European Conference on Computer Vision (ECCV), 2014.

\bibitem{yosinski_2014}
J.~Yosinski, J.~Clune, Y.~Bengio, H.~Lipson, How transferable are features in
  deep neural networks?, in: Advances in Neural Information Processing Systems
  (NIPS), 2014.

\bibitem{chatfield_2014}
K.~Chatfield, K.~Simonyan, A.~Vedaldi, A.~Zisserman, Return of the devil in the
  details: {D}elving deep into convolutional nets, in: British Machine Vision
  Conference (BMVC), 2014.

\bibitem{torralba_2011}
A.~Torralba, A.~A. Efros, Unbiased look at dataset bias, in: IEEE Conference on
  Computer Vision and Pattern Recognition (CVPR), 2011.

\bibitem{graves_2016_act}
A.~Graves, Adaptive computation time for recurrent neural networks,
  arXiv:1603.08983.

\bibitem{zhou_2014}
B.~Zhou, A.~Lapedriza, J.~Xiao, A.~Torralba, A.~Oliva, Learning deep features
  for scene recognition using places database, in: Advances in Neural
  Information Processing Systems (NIPS), 2014.

\bibitem{long_2015}
J.~Long, E.~Shelhamer, T.~Darrell, Fully convolutional networks for semantic
  segmentation, in: IEEE Conference on Computer Vision and Pattern Recognition
  (CVPR), 2015.

\bibitem{tang_2013}
Y.~Tang, Deep learning using linear support vector machines, in: International
  Conference on Machine Learning Workshop (ICMLW) on Challenges in
  Representation Learning, 2013.

\bibitem{szegedy_2014_googlenet}
C.~Szegedy, W.~Liu, Y.~Jia, P.~Sermanet, S.~Reed, D.~Anguelov, D.~Erhan,
  V.~Vanhoucke, A.~Rabinovich, Going deeper with convolutions, in: IEEE
  Conference on Computer Vision and Pattern Recognition (CVPR), 2015.

\bibitem{he_2015_resnet}
K.~He, X.~Zhang, S.~Ren, J.~Sun, Deep residual learning for image recognition,
  in: IEEE Conference on Computer Vision and Pattern Recognition (CVPR), 2016.

\end{thebibliography}

\end{document}